\documentclass{ecai}
\usepackage{times}
\usepackage{graphicx}
\usepackage{latexsym}

\usepackage{amsfonts} 
\usepackage{amsmath} 
\usepackage{subfigure}
\usepackage{multirow}
\usepackage{appendix}

\begin{document}

\title{Adaptive Interaction Fusion Networks for Fake News Detection}

\author{Lianwei Wu \and Yuan Rao$^*$\institute{1) Lab of Social Intelligence and Complex Data Processing, Software School, Xi'an Jiaotong University, Xi'an, China; 2) Shannxi Joint Key Laboratory for Artifact Intelligence (Sub-Lab of Xi'an Jiaotong University), Xi'an, China; 3) Research Institute of Xi'an Jiaotong University, Shenzhen, China, email: stayhungry@stu.xjtu.edu.cn; raoyuan@mail.xjtu.edu.cn \newline*Corresponding author}}

\maketitle
\bibliographystyle{ecai}

\begin{abstract}
  The majority of existing methods for fake news detection universally focus on learning and fusing various features for detection. However, the learning of various features is independent, which leads to a lack of cross-interaction fusion between features on social media, especially between posts and comments. Generally, in fake news, there are emotional associations and semantic conflicts between posts and comments. How to represent and fuse the cross-interaction between both is a key challenge. In this paper, we propose Adaptive Interaction Fusion Networks (AIFN) to fulfill cross-interaction fusion among features for fake news detection. In AIFN, to discover semantic conflicts, we design gated adaptive interaction networks (GAIN) to capture adaptively similar semantics and conflicting semantics between posts and comments. To establish feature associations, we devise semantic-level fusion self-attention networks (SFSN) to enhance semantic correlations and fusion among features. Extensive experiments on two real-world datasets, i.e., RumourEval and PHEME, demonstrate that AIFN achieves the state-of-the-art performance and boosts accuracy by more than 2.05\% and 1.90\%, respectively.
\end{abstract}

\section{Introduction}
Owing to the low cost of information dissemination and the lack of timely and effective supervision, social media provides an ideal breeding ground for the growth of fake news (a.k.a. hoaxes, rumors, etc.). Research indicates that fake news has dominated the news cycle since the US presidential election (2016) \cite{allcott2017social,grinberg2019fake}. In detail, a tweeting rate for users tweeting links to websites containing news classified as fake more than four times larger than for traditional media and 25\% of tweets during the election in Twitter spread fake news \cite{bovet2019influence}, meanwhile, 1\% of users are even exposed to 80\% of fake news \cite{grinberg2019fake}. Furthermore, fake news has greater virality than true information, which diffuses remarkably farther, faster, deeper, and more broadly than the truth \cite{lazer2018science}. Therefore, how to effectively evaluate information credibility has become a crucial problem, drawing wide attention from academic and industrial communities.

\begin{figure}
	\centering
	\includegraphics[width=0.47\textwidth]{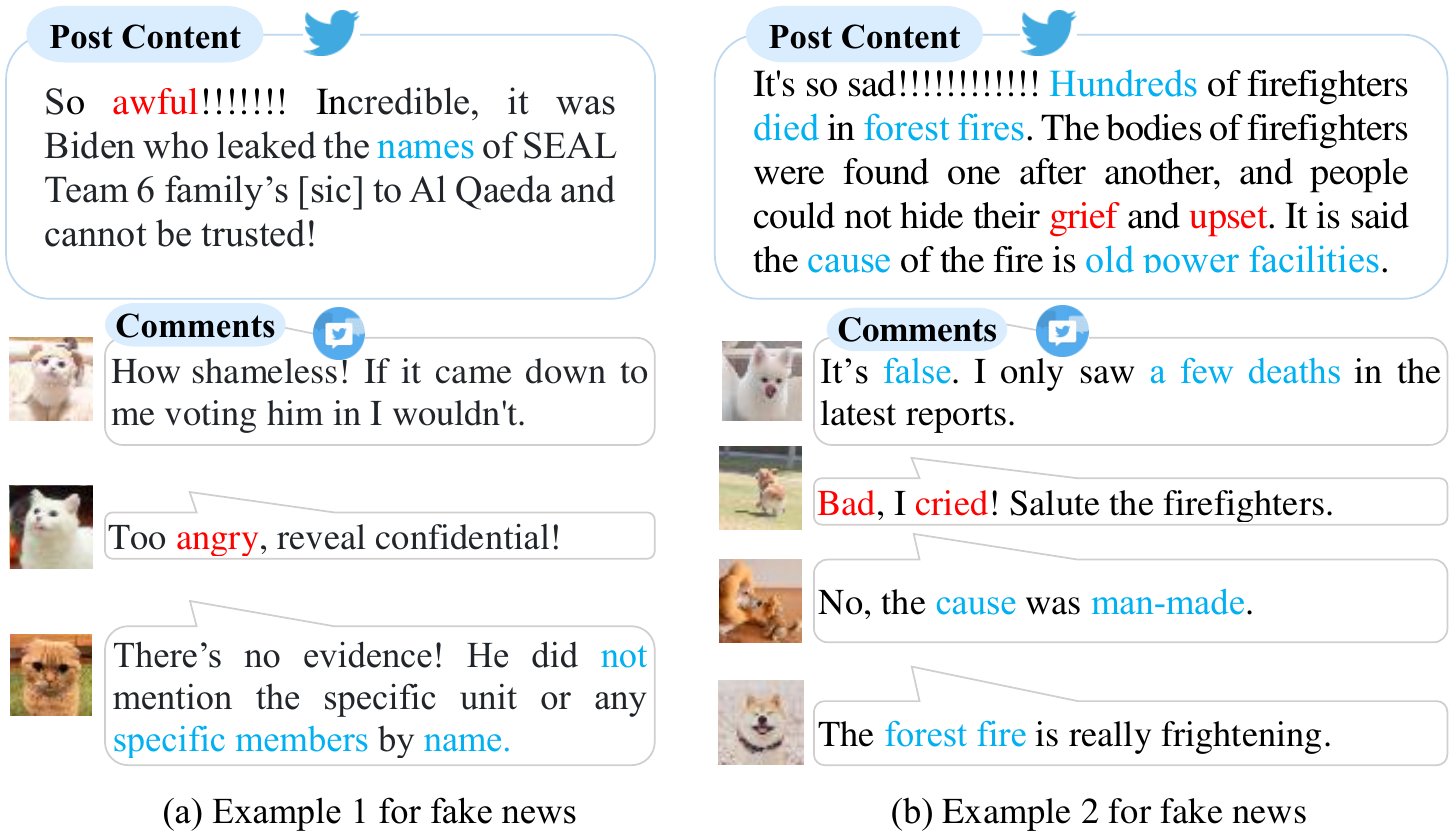}
	\caption{Two fake news from Twitter.\hspace{-0.07mm} Blue denotes semantic correlations between comments and posts while red means emotional correlations between comments and posts.} \label{Fig1examples}
\end{figure}

Currently, most existing studies based on deep neural networks capture various features for fake news detection, which have achieved great success. Generally, they devise different modules to capture different credibility features including news content \cite{khattar2019mvae,ajao2019sentiment}, diffusion patterns \cite{ma2018rumor,liu2018early}, and social context (a.k.a. meta-data) \cite{zhang2019reply,yang2019xfake}, and then use some traditional fusion strategies to integrate these features for evaluation. For instance, Wang \cite{wang2017liar} adopts CNN to capture content semantics and employs BiLSTM to learn meta-data features. Finally, the author fuses them by concatenation for fake news detection. Zhang et al. \cite{zhang2019reply} obtain semantics of news and comments by different BiLSTM combining Bayesian networks then build simple multi-layer perception to fuse them for misinformation detection. Wu et al. \cite{wu2020discovering} rely on adversarial networks and multi-task learning to capture differential credibility features from news content semantics and fuse them for information credibility evaluation.

However, there are still several general limitations of the above methods. \textbf{First}, feature fusion is relatively shallow. Most fusion strategies are based on concatenation, addition, or simple neural networks, which are hard to choose noteworthy features and even bring a certain amount of noise. \textbf{Secondly}, various features are typically fused only in the final evaluation stage while lacking cross-interaction fusion in the intermediate semantic learning stage. We know that cross interaction is ubiquitous on social media, especially there are feature correlations and semantic conflicts between posts and comments in fake news. Figure \ref{Fig1examples} shows two concrete examples to illustrate this phenomenon. We observe that: 1) Emotions of comments usually correspond to these of posts (in red), like `cried' and `grief'; and 2) Some comments have semantic conflicts with posts (in blue), like `power facilities' and `man-made'. Cross-interaction fusion among features can enhance core semantics and help to discover the causes of fake news. Therefore, how to represent and fuse these interactions is a major challenge for fake news detection.

To address the above problems, we propose \textbf{A}daptive \textbf{I}nteraction \textbf{F}usion \textbf{N}etworks (henceforth, AIFN) to cross-interactively fuse various features for fake news detection. Specifically, in order to discover semantic conflicts between posts and comments, we design gated adaptive interaction networks (GAIN) to acquire adaptively similar and conflicting semantics between both, in which conflicting gate is designed for capturing differential features while refining gate aims to obtain similar semantics between posts and comments. To effectively utilize feature associations within posts and comments respectively, we explore semantic-level fusion self-attention networks (SFSN) to screen their valuable features and fuse them deeply. Experimental results show that AIFN achieves better performance than other state-of-the-art methods and gains new benchmarks. The main contributions of our work can be summarized as follows:

\begin{itemize}
\item Proposed gated adaptive interaction networks fulfill global feature interaction, which focuses on discovering conflicting semantics more accurately by capturing adaptively similar semantics and differential semantics (Section \ref{balancedEval} and \ref{caseStudy}).
\item Explored semantic-level fusion self-attention networks can achieve deep semantic fusion within posts and comments, which effectively build feature associations between both (Section \ref{multiScaleEval} and Section \ref{caseStudy}).
\item Experiments on two public, widely used fake news datasets demonstrate that our method significantly outperforms previous state-of-the-art methods (Section \ref{perfEval}). We release the source code publicly for further research\footnote{https://www.dropbox.com/s/9pos1pvv392adnt/interactFusions.zip?dl=0}.
\end{itemize}

The rest of the paper is organized as follows. We start with an overview of related work in Section \ref{sec2related}. Section \ref{sec3method} presents the details of our approach. Experiment results and discussions are given in Section \ref{sec4experiment}, and Section \ref{sec5conclusion} concludes our work.

\section{Related Work}
\label{sec2related}
\subsection{Fake News Detection}
Existing studies for fake news detection can be roughly summarized into two categories. The first category is to extract or construct plenty of features with manual ways \cite{castillo2011information,zhao2015enquiring}. The prominent merits of this category are low computational complexity and many effective meta-data features discovered from social context.

Instead of gaining features by labor-intensive manual design, the second category is to automatically capture credibility features based on deep neural networks. There are two ways: One is to capture linguistic features from text content, such as semantics \cite{wu2018false,qian18,ma2019detect}, emotions \cite{ajao2019sentiment}, writing styles \cite{potthast2018stylometric}, and stances \cite{wu2019different}. But the capture of stances might bring some noise and reduce some performance indicators of models. The other way emphasizes on capturing meta-data features surrounding social context such as source-based \cite{tschiatschek2018fake}, post-based \cite{lianwei28multi,zhang2019reply}, comment-based \cite{shu2019defend}, user-based \cite{shu2019role}, and propagation-based \cite{ma2018rumor}. In this work, combined with the pros of both categories, we capture deeply semantic and emotional features from posts and comments.
\subsection{Feature Fusion}
Generally, most feature fusion methods are based on concatenation \cite{nguyen2019fake}, point-wise addition \cite{liang2019multi}, and multi-layer perceptions \cite{zhang2019reply}, etc. For example, Nguyen et al. \cite{nguyen2019fake} concatenate document features obtained by TF-IDF, semantics obtained by word2vec, and user-event graph features obtained by node2vec as the input layer of the model for fake news detection. Nevertheless, these fusion methods usually lack feature interaction. To address it, attention mechanism \cite{shu2019role,zhang2019interactive} is developed to focus on the interaction of two related features, like user profiles and behaviors, posts and comments. Typically, Shu et al. \cite{shu2019defend} present a sentence-comment co-attention network to exploit both news contents and comments to jointly detect explainable check-worthy sentences and comments for fake news detection. But attention-based methods are mostly based on the interaction of local and partial features, which lack global cross-interaction of all features. To address these limitations, we propose cross-interaction fusion model integrating self-attention networks for evaluation.

\begin{figure*}
	\centering
	\includegraphics[width=1\textwidth]{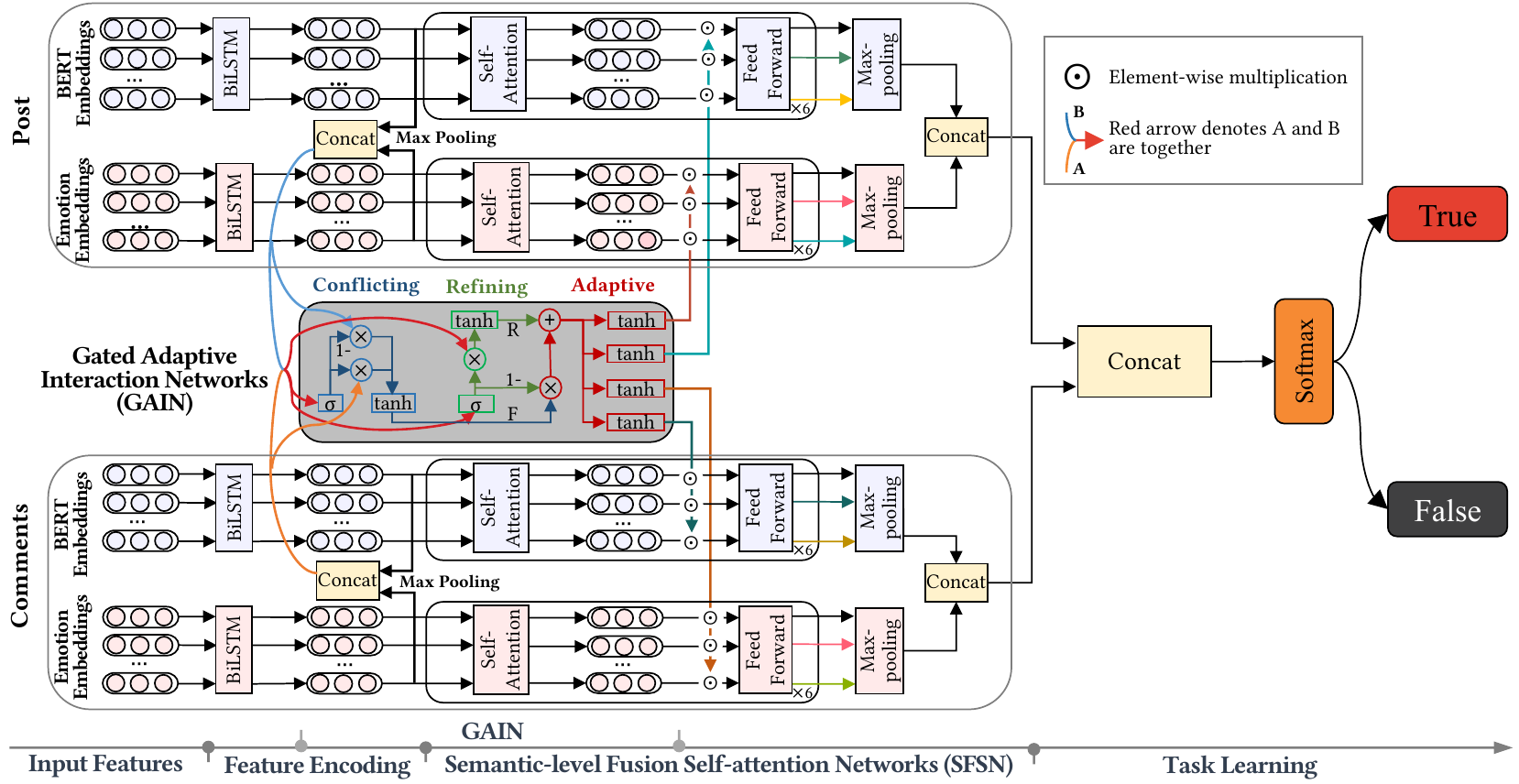}
	\caption{Overall architecture of AIFN. The model utilizing organically four types of features focuses on interaction and fusion among features by the following components: input features, feature encoding, semantic-level fusion self-attention networks, gated adaptive interaction networks, and task learning.}
	\label{Fig2model}
\end{figure*}

\section{Adaptive Interaction Fusion Networks}
\label{sec3method}
The architecture of AIFN with five components is illustrated in Figure \ref{Fig2model} and each component of AIFN is described as follows:

\subsection{Input Feature Descriptions}
AIFN learns four types of features around posts and comments from the perspectives of word and emotion.

\subsubsection{BERT Embeddings}
Word embeddings of posts and comments in tweets are both applied BERT embeddings. BERT embeddings of a post containing $l$ words are indicated as ${\rm \textbf{X}}^p= \{[{\rm \textbf{x}}_1^p; {\rm \textbf{pos}}_1^p ],[{\rm \textbf{x}}_2^p; {\rm \textbf{pos}}_2^p ], ..., [{\rm \textbf{x}}_l^p; {\rm \textbf{pos}}_l^p]\}$, ${\rm \textbf{X}}^p \in \mathbb{R}^{l \times (d+l)}$, where embeddings of each word are concatenated by word embeddings ${\rm \textbf{x}}_i^p \in \mathbb{R}^d$ and position embeddings ${\rm \textbf{pos}}_i^p \in \mathbb{R}^l$. ${\rm \textbf{x}}_i^p$ of word $i$ is a d-dimensional vector obtained by pre-trained BERT model \cite{devlin2019bert}. ${\rm \textbf{pos}}_i^p$ of word $i$ is achieved by one-hot encoding. $;$ means concatenation operation. For BERT embeddings of comments, we first rank all comments in chronological order and then concatenate them into a sequence with $k$ words. The way of embeddings is the same as the post, i.e., ${\rm \textbf{X}}^c={[{\rm \textbf{x}}_1^c; {\rm \textbf{pos}}_1^c],[{\rm \textbf{x}}_2^c; {\rm \textbf{pos}}_2^c],  ..., [{\rm \textbf{x}}_k^c; {\rm \textbf{pos}}_k^c]}$, ${\rm \textbf{X}}^c \in \mathbb{R}^{k \times (d+k)}$.

\subsubsection{Emotion Embeddings}
We represent D-dimensional emotion embeddings of one word via pre-trained emotion word embeddings (EWE) \cite{agrawal2018learning}, which learns emotion-enriched word representations by projecting emotionally similar words into neighboring spaces based on a large training dataset with a rich spectrum of emotions. Here, emotion embeddings of a post ${\rm \textbf{E}}^p$ and comments ${\rm \textbf{E}}^c$ are denoted as ${\rm \textbf{E}}^p=\{{\rm \textbf{e}}_1^p, {\rm \textbf{e}}_2^p, ..., {\rm \textbf{e}}_l^p\}$ and ${\rm \textbf{E}}^c=\{{\rm \textbf{e}}_1^c, {\rm \textbf{e}}_2^c, ..., {\rm \textbf{e}}_k^c\}$, respectively, where ${\rm \textbf{e}}_i^p \in \mathbb{R}^D$, ${\rm \textbf{E}}^p \in \mathbb{R}^{l \times D}$, ${\rm \textbf{e}}_i^c \in \mathbb{R}^D$ and ${\rm \textbf{E}}^c \in \mathbb{R}^{k \times D}$.

\subsection{Feature Encoding}
We encode word-based and emotion-based sequence via BiLSTM, which can not only capture long-term dependencies by a persistent memory compared with RNN but also obtain contextual information by modeling word sequence from both directions of words. BiLSTM contains a forward LSTM $\overrightarrow{{\rm \textbf{LSTM}}} $ and a backward LSTM $ \overleftarrow{{\rm \textbf{LSTM}}}$, which learns word sequence from the first word to the last word and learns the sequence in reverse order respectively.

\begin{eqnarray}\label{eq1}
\overrightarrow{{\rm \textbf{h}}_i} &=&\overrightarrow{{\rm \textbf{LSTM}}}(\overrightarrow{{\rm \textbf{h}}_{i-1}},{\rm \textbf{x}}_i) \\
\overleftarrow{{\rm \textbf{h}}_i} &=&\overleftarrow{{\rm \textbf{LSTM}}}(\overleftarrow{{\rm \textbf{h}}_{i+1}},{\rm \textbf{x}}_i) \\
{\rm \textbf{h}}_i &=&\overrightarrow{{\rm \textbf{h}}_i} \oplus \overleftarrow{{\rm \textbf{h}}_i}
\end{eqnarray}
where $\overrightarrow{{\rm \textbf{h}}_i} \in \mathbb{R}^h$ and $\overleftarrow{{\rm \textbf{h}}_i} \in \mathbb{R}^h$ are hidden states of forward and backward LSTM at position $i$, respectively. ${\rm \textbf{x}}_i$ is the $i$-th input which can be replaced by word embeddings ${\rm \textbf{X}}_i^p$ and ${\rm \textbf{X}}_i^c$, or emotion embeddings ${\rm \textbf{E}}_i^p$ and ${\rm \textbf{E}}_i^c$. $\oplus$ denotes concatenation. In addition, experiments confirm BiLSTM can be replaced by BiGRU for comparable performance in AIFN.

\subsection{Gated Adaptive Interaction Networks (GAIN)}
\label{sub-strategy}
We first obtain the pooled vectors ${\rm \textbf{X}}_{pool}^{pw}$, ${\rm \textbf{X}}_{pool}^{pe}$, ${\rm \textbf{X}}_{pool}^{cw}$, and ${\rm \textbf{X}}_{pool}^{ce}$ (all $\in \mathbb{R}^l$) from the encoded word-level and the encoded emotion-level features in posts and comments respectively by max-pooling operation. Then we get the integrated post vector ${\rm \textbf{X}}_{pool}^p$ and comment vector ${\rm \textbf{X}}_{pool}^c$ by concatenation.

\begin{eqnarray}\label{eq201}
{\rm \textbf{X}}_{pool}^p &=& {\rm \textbf{X}}_{pool}^{pw} \oplus {\rm \textbf{X}}_{pool}^{pe} \\
{\rm \textbf{X}}_{pool}^c &=& {\rm \textbf{X}}_{pool}^{cw} \oplus {\rm \textbf{X}}_{pool}^{ce}
\end{eqnarray}

To discover semantic conflicts between posts and comments, we devise GAIN to capture adaptively similar semantics and differential semantics.
\subsubsection{Conflicting Gate}

Conflicting gate is to obtain differential and conflicting features between posts and comments, i.e., ${\rm \textbf{X}}_{pool}^p$ and ${\rm \textbf{X}}_{pool}^c$, respectively.

\begin{eqnarray}\label{eq20}
\mu_f &=& \sigma({\rm \textbf{X}}_{pool}^p {\rm \textbf{W}}_{f1}+{\rm \textbf{X}}_{pool}^c {\rm \textbf{W}}_{f2}+{\rm \textbf{b}}_f) \\
{\rm \textbf{F}}\! &=& \! {\rm \textbf{tanh}}⁡({\rm \textbf{X}}_{pool}^p\! \odot\! \mu_f {\rm \textbf{W}}_{h1}\!+\!{\rm \textbf{X}}_{pool}^c \odot(1\!-\!\mu_f){\rm \textbf{W}}_{h2}\!+\!{\rm \textbf{b}}_h)
\end{eqnarray}
where ${\rm \textbf{F}}$ is the conflicting features obtained. All weights ${\rm \textbf{W}}\in R^{2l\times 2l}$ and biases ${\rm \textbf{b}}_f, {\rm \textbf{b}}_h \in R^{2l}$.

When $\mu_f$ approaches 0, the conflicting features in ${\rm \textbf{X}}_{pool}^p$ are more obtained. Conversely, more conflicting features in ${\rm \textbf{X}}_{pool}^c$ are obtained. This way ensures the adjustability of the conflicting gate so that all conflicting features of both have the possibility of being obtained.

\subsubsection{Refining Gate}
We design refining gate to capture more similar features between posts and comments ${\rm \textbf{X}}_{pool}^p$ and ${\rm \textbf{X}}_{pool}^c$:

\begin{eqnarray}\label{eq22}
\mu_r &=& \sigma({\rm \textbf{X}}_{pool}^p {\rm \textbf{W}}_{r1}+{\rm \textbf{X}}_{pool}^c {\rm \textbf{W}}_{r2}+{\rm \textbf{b}}_r) \\
{\rm \textbf{R}} &=& {\rm \textbf{tanh}}⁡({\rm \textbf{X}}_{pool}^p \odot \mu_r {\rm \textbf{W}}_{rp} + {\rm \textbf{X}}_{pool}^c \odot \mu_r  {\rm \textbf{W}}_{rc}+{\rm \textbf{b}}_{rr})
\end{eqnarray}
where all weights ${\rm \textbf{W}}\in R^{2l\times 2l}$ and biases ${\rm \textbf{b}}\in R^{2l}$.

\subsubsection{Adaptive Mechanism}
The location of semantic conflicts is inseparable from the matching of highly similar semantics in sequences, in order to accurately locate the differential features in a sequence, we develop adaptive mechanism to capture adaptively the similar and conflicting semantics.

\begin{eqnarray}\label{eq24}
{\rm \textbf{S}} &=& {\rm \textbf{R}}+(1-\mu_r)\odot {\rm \textbf{F}} \\
{\rm \textbf{t}}^{px} &=& {\rm \textbf{tanh}}({\rm \textbf{W}}^{px}{\rm \textbf{S}}+{\rm \textbf{b}}^{px}) \\
{\rm \textbf{t}}^{pe} &=& {\rm \textbf{tanh}}({\rm \textbf{W}}^{pe}{\rm \textbf{S}}+{\rm \textbf{b}}^{pe}) \\
{\rm \textbf{t}}^{cx} &=& {\rm \textbf{tanh}}({\rm \textbf{W}}^{cx}{\rm \textbf{S}}+{\rm \textbf{b}}^{cx}) \\
{\rm \textbf{t}}^{ce} &=& {\rm \textbf{tanh}}({\rm \textbf{W}}^{ce}{\rm \textbf{S}}+{\rm \textbf{b}}^{ce})
\end{eqnarray}
where ${\rm \textbf{S}} \in \mathbb{R}^{d_x}$ is the adaptive features. ${\rm \textbf{t}}^{px}$, ${\rm \textbf{t}}^{pe}$, ${\rm \textbf{t}}^{cx}$ and ${\rm \textbf{t}}^{ce}$ are different dimensional global interaction vectors suitable for different types of features, i.e., word-level in posts, emotion-level in posts, word-level in comments, and emotion-level in comments, respectively.

\subsection{Semantic-level Fusion Self-attention Networks (SFSN)}
We design SFSN to build  associations and deep semantic fusion among features. Specifically, SFSN is used to fuse deeply different types of features and the interacted features obtained by GAIN (in Section \ref{sub-strategy}). We take word-level in posts as an example to explain the details, where ${\rm \textbf{t}}^{px}$ is the global interaction vector and ${\rm \textbf{X}}^{px}=\{{\rm \textbf{x}}_1^{px}, {\rm \textbf{x}}_2^{px}, ..., {\rm \textbf{x}}_l^{px}\}$ is outputs of BiLSTM for word-level in posts.

\textbf{Self-attention Networks } We leverage the multi-head self-attention networks to learn dependencies and semantics between any two words in one sequence. The scaled dot-product attention, the core of self-attention networks, is described as

\begin{equation}\label{eq14}
{\rm \textbf{Attention}}({\rm \textbf{Q}}, {\rm \textbf{K}}, {\rm \textbf{V}}) = {\rm \textbf{softmax}}(\frac{{\rm \textbf{Q}}{\rm \textbf{K}}^T}{\sqrt{d}}){\rm \textbf{V}}
\end{equation}
where ${\rm \textbf{Q}} \in \mathbb{R}^{l\times 2h}$, ${\rm \textbf{K}} \in \mathbb{R}^{l\times 2h}$, and ${\rm \textbf{V}} \in \mathbb{R}^{l\times 2h}$ are query, key, and value matrices respectively and ${\rm \textbf{Q}}={\rm \textbf{K}}={\rm \textbf{V}}={\rm \textbf{X}}^{px}$.

To get high parallelizability of attention, multi-head attention first linearly projects queries, keys, and values $j$ times by different linear projections and then $j$ projections perform the scaled dot-product attention in parallel. Finally, these results of attention are concatenated and once again projected to get the new representation. Formally, the multi-head attention can be formulated as:

\begin{eqnarray}\label{eq15}
{\rm head}_i&=&{\rm \textbf{Attention}}({\rm \textbf{Q}}{\rm \textbf{W}}_i^Q, {\rm \textbf{K}}{\rm \textbf{W}}_i^K, {\rm \textbf{V}}{\rm \textbf{W}}_i^V) \\
{\rm \textbf{O}}^\prime &=& {\rm \textbf{MultiHead}}({\rm \textbf{Q}}, {\rm \textbf{K}}, {\rm \textbf{V}}) \\
&=& {\rm \textbf{Concat}}({\rm head}_1, {\rm head}_2, ..., {\rm head}_h){\rm \textbf{W}}^o
\end{eqnarray}
where ${\rm \textbf{W}}_i^Q\in \mathbb{R}^{2h\times d_k}$, ${\rm \textbf{W}}_i^K\in \mathbb{R}^{2h\times d_k}$, ${\rm \textbf{W}}_i^V\in \mathbb{R}^{2h\times d_k}$, and ${\rm \textbf{W}}^o\in \mathbb{R}^{2h\times 2h}$ are trainable parameters and $d_k$ is $2h/j$.

\textbf{Fusion Strategy} We exploit the interaction vector ${\rm \textbf{t}}^{px}$ to concentrate on the fusion of deep semantics in self-attention networks for highlighting semantic relevance of different types of features.

\begin{equation}\label{eq17}
{\rm \textbf{O}}^{px}={\rm \textbf{t}}^{px} \odot {\rm \textbf{O}}^\prime
\end{equation}
where $\odot$ is element-wise multiplication.

Subsequently, self-attention networks pass a feed forward network (FFN) for adding non-linear features while keeping scale-invariant features, which includes a single hidden layer with an ReLU.

\begin{equation}\label{eq18}
{\rm \textbf{O}}_{ffn}^{px}={\rm \textbf{FFN}}({\rm \textbf{O}}^{px})={\rm \textbf{max}}(0, {\rm \textbf{O}}^{px}{\rm \textbf{W}}_1+b_1){\rm \textbf{W}}_2+{\rm \textbf{b}}_2
\end{equation}
where ${\rm \textbf{W}}_1$, and ${\rm \textbf{W}}_2$ (both $\in \mathbb{R}^{2h\times 2h}$), ${\rm \textbf{b}}_1$, and ${\rm \textbf{b}}_2$ (both $\in \mathbb{R}^{2h}$) are learned parameters.

Finally, we apply max-pooling to each dimension across all words for gaining a fixed-size representation ${\rm \textbf{X}}^{pw}$ of the sequence as final outputs of word-level in posts.

\begin{equation}\label{eq19}
{\rm \textbf{X}}^{pw}={\rm \textbf{max-pooling}}({\rm \textbf{O}}_{ffn}^{px})
\end{equation}

Additionally, we leverage concatenation to integrate the learned word-level features and emotion-level features from posts and comments, respectively, i.e., ${\rm \textbf{X}}^{pw}$, ${\rm \textbf{X}}^{pe}$, ${\rm \textbf{X}}^{cw}$, and ${\rm \textbf{X}}^{ce}$.

\begin{eqnarray}\label{eq202}
{\rm \textbf{X}}^p &=& {\rm \textbf{X}}^{pw} \oplus {\rm \textbf{X}}^{pe} \\
{\rm \textbf{X}}^c &=& {\rm \textbf{X}}^{cw} \oplus {\rm \textbf{X}}^{ce}
\end{eqnarray}
where ${\rm \textbf{X}}^p$ and ${\rm \textbf{X}}^c$ are the final integrated features from posts and comments respectively.

\subsection{Task Learning}
We employ concatenation operation to integrate the output features of posts and comments, i.e., ${\rm \textbf{X}}^p$ and ${\rm \textbf{X}}^c$.
\begin{equation}\label{eq202}
{\rm \textbf{X}}^{pc} = {\rm \textbf{X}}^p \oplus {\rm \textbf{X}}^c
\end{equation}

Then we apply the integrated features to Softmax function for task learning. Softmax function emits the prediction of probability distribution by the following equations:

\begin{equation}\label{eq29}
{\rm \textbf{p}}={\rm \textbf{softmax}}({\rm \textbf{W}}_d {\rm \textbf{X}}^{pc}+{\rm \textbf{b}}_d)
\end{equation}

We train the model to minimize cross-entropy error for a single training instance with ground-truth label ${\rm \textbf{y}}$:

\begin{equation}\label{eq30}
{\rm \textbf{Loss}}=-\sum {\rm \textbf{y}}log{\rm \textbf{p}}
\end{equation}

\section{Experiments}
\label{sec4experiment}

\subsection{Datasets and Evaluation Metrics}
For experimental evaluation, we use two real-world benchmark datasets, RumourEval \cite{derczynski2017semeval} and PHEME \cite{zubiaga2016analysing}, which respectively contains 325 and 6,425 Twitter threads discussing rumors. Both datasets include Twitter conversation threads associated with different newsworthy events, like the Ferguson unrest, where one thread consists of a source tweet conveying a rumor and a tree of comments. The credibility of each tweet can be true, false, and unverified. Since our goal is to evaluate whether one tweet is true or false, we filter out unverified tweets. Table \ref{tab1statisDatasets} gives statistics of the two datasets.

In consideration of the imbalance label distributions, evaluation solely relying on accuracy effortlessly achieve competitive performance beyond the majority class. Therefore, besides accuracy (A), we add precision (P), recall (R) and F1-score (F1) as complementary evaluation metrics for tasks. We divide the two datasets into training, validation, and testing subsets with proportion of 70\%, 10\%, and 20\%.

\begin{table}
\footnotesize
	\center
	\caption{Statistics of the datasets}
	\label{tab1statisDatasets}
	\begin{tabular}{cccccc} \hline
        \rule{0pt}{12pt}
		Subset & Veracity &\multicolumn{2}{c}{RumourEval} &\multicolumn{2}{c}{PHEME} \\
        \cline{3-4} \cline{5-6} \rule{0pt}{12pt}
		& & \#posts & \#comments & \#posts & \#comments \\ \hline
        \\[-6pt]
		\multirow{3}*{Training} & True & 83 & 1,949 & 861 & 24,438 \\
		& False & 70 & 1,504 & 625 & 17,676 \\
		& Total & 153 & 3,453 & 1,468 & 42,114 \\ \\[-6pt]
		\multirow{3}*{Validation} & True & 10 & 101 & 95 & 1,154 \\
		& False & 12 & 141 & 115 & 1,611 \\
		& Total & 22 & 242 & 210 & 2,765 \\ \\[-6pt]
		\multirow{3}*{Testing} & True & 9 & 412 & 198 & 3,077 \\
		& False & 12 & 437 & 219 & 3,265 \\
		& Total & 21 & 849 & 417 & 6,342 \\ \hline\\
	\end{tabular}
\end{table}

\subsection{Setting}
We strictly turn all hyper-parameters on the validation dataset, and we achieve the best performance via a small grid search. The hyper-parameters turned on the validation subset are shown as follows:

\begin{itemize}
\item Word embedding sizes $d$ for posts and comments are both set to 768;
\item Emotion embeddings size $D$ is 300;
\item The dimensionality of LSTM hidden state $h$ is 120;
\item The initial learning rate is set to 0.001;
\item The dropout rate is 0.4;
\item The number of projections $j$ is 8;
\item Attention heads and blocks are set to 6 and 4, respectively;
\item The minibatch size is 64.
\end{itemize}

\begin{table*}
\begin{center}
\footnotesize
	\caption{Performance comparison of our proposed model against the baselines.}
	\label{Tab2performEval}
\setlength{\tabcolsep}{4.07mm}{
	\begin{tabular}{llcccccccc}
		\hline \rule{0pt}{12pt}
		Dataset & Measure & SVM & CNN & TE & DeClarE & MTL-LSTM & TRNN & Bayesian-DL & Ours \\ \hline
        \\[-6pt]
		\multirow{4}*{RumourEval} & A (\%) & 71.42 & 61.90 & 66.67 & 66.67 & 66.67 & 76.19 & 80.95 & \textbf{82.89} \\
		& P (\%) & 66.67 & 54.54 & 60.00 & 58.33 & 57.14 & 70.00 & 77.78 & \textbf{78.52} \\
		& R (\%) & 66.67 & 66.67 & 66.67 & 77.78 & \textbf{88.89} & 77.78 & 77.78 & 86.21 \\
		& F1 (\%) & 66.67 & 59.88 & 63.15 & 66.67 & 69.57 & 73.68 & 77.78 & \textbf{82.19} \\ \\[-6pt]
		\multirow{4}*{PHEME} & A (\%) & 72.18 & 59.23 & 65.22 & 67.87 & 74.94 & 78.65 & 80.33 & \textbf{82.45} \\
		& P (\%) & 78.80 & 56.14 & 63.05 & 64.68 & 68.77 & 77.11 & 78.29 & \textbf{79.12} \\
		& R (\%) & 75.75 & 64.64 & 64.64 & 71.21 & \textbf{87.87} & 78.28 & 79.29 & 85.20 \\
		& F1 (\%) & 72.10 & 60.09 & 63.83 & 67.89 & 77.15 & 77.69 & 78.78 & \textbf{82.05} \\
		\hline
        \\[-6pt]
	\end{tabular}
}
\end{center}
\end{table*}

\subsection{Performance Evaluation}
\label{perfEval}
We compare AIFN with the following baselines:

\textbf{SVM} \cite{derczynski2017semeval} are used to detect misinformation based on manually extracted features.

\textbf{CNN} \cite{chen2017ikm} with different convolutional window sizes captures content features similar to n-grams for rumor verification.

\textbf{TE} \cite{guacho2018semi} leverages tensor decomposition to derive concise article embeddings to create an article-by-article graph for misinformation detection.

\textbf{DeClarE} \cite{popat2018declare} applies posts as the queries of attention mechanism to extract evidence from comments and aggregates post sources and languages.

\textbf{MTL-LSTM} \cite{kochkina2018all} jointly trains three tasks, i.e., rumor detection, veracity classification, and stance classification, and learns the relationships among the tasks by common LSTM for improving performance of each task.

\textbf{TRNN} \cite{ma2018rumor} contains two tree-structured models based on RNN, which learns credibility features based on non-sequential propagation structure formed by comments for rumor detection. In this work, we adopt the top-down model with better results as the baseline.

\textbf{Bayesian-DL} \cite{zhang2019reply} employs Bayesian to represent the uncertainty of prediction for the veracity of a claim and then encodes all the people's comments to the claim through LSTM for misinformation detection.

The experimental results are summarized in Table \ref{Tab2performEval}. We observe that:
\begin{itemize}
\item MTL-LSTM exploiting stance features shows 2.86\% and 2.67\% improvements in recall on RumourEval and PHEME respectively, but it also introduces noise, which makes it achieve from 4.90\% to 21.38\% degradation than our model in accuracy, precision, and F1. Besides, our model achieves 1.94\% and 2.12\% boosts than the latest baseline (Bayesian-DL) in accuracy on the two datasets respectively. These reveal the effectiveness of our model.
\item SVM incorporating meta-data features gains better performance than some deep neural networks only learning content features, like CNN and TE. It confirms the effectiveness of meta-data features for fake news detection.
\item Both TRNN and Bayesian-DL by integrating comments into posts achieves eminent performance than other baselines, which reveals the effectiveness of the fusion of comments and posts. But they are worse than AIFN because AIFN not only captures semantics of comments and posts but also emotional features from the both.
\end{itemize}

\begin{table*}
\footnotesize
	\center
	\caption{Comparison between AIFN and simplified models on RumourEval and PHEME.}
	\label{tab3ablation}
\setlength{\tabcolsep}{5.05mm}{
	\begin{tabular}{lcccccccc} \hline
\rule{0pt}{12pt}
		&\multicolumn{4}{c}{RumourEval} &\multicolumn{4}{c}{PHEME} \\ \cline{2-9}
\rule{0pt}{12pt}
		&A(\%) & P(\%) & R(\%) & F1(\%) &A(\%) & P(\%) & R(\%) & F1(\%) \\ \hline
        \\[-6pt]
		BERT+BiLSTM &75.54 & 70.22 & 75.26 & 72.65 & 75.30 & 73.89 & 76.21 & 75.03 \\
		BERT+BiLSTM+SFSN & 80.24 & 77.18 & 84.23 & 80.79 & 80.96 & 77.56 & 83.66 & 80.49 \\
		BERT+BiLSTM+SFSN+GAIN & 82.89 & 78.52 & 86.21 & 82.19 & 82.45 & 79.12 & 85.20 & 82.05 \\
        Glove+BiLSTM+SFSN+GAIN & 80.04 & 77.21 & 84.90 & 80.87 & 80.80 & 77.13 & 83.75 & 80.30 \\ \hline
        \\[-6pt]
	\end{tabular}
}
\end{table*}

\subsection{Discussions}
We evaluate the effectiveness of each part of AIFN and the effectiveness of internal structure of GAIN and SFSN.
\subsubsection{Ablation Analysis of AIFN}
Table \ref{tab3ablation} provides the experimental results of AIFN and the following simplified models:
\begin{itemize}
\item \textbf{BERT+BiLSTM} applies BERT to BiLSTM model.
\item \textbf{BERT+BiLSTM+SFSN} adds semantic-level fusion self-attention networks (SFSN) to BERT+BiLSTM.
\item \textbf{BERT+BiLSTM+SFSN+GAIN} means BERT+BiLSTM+SFSN incorporating gated adaptive interaction networks (GAIN), i.e., AIFN.
\item \textbf{Glove+BiLSTM+SFSN+GAIN} uses Glove \cite{pennington2014glove} as word embeddings of AIFN.
\end{itemize}
From the experimental results of Table \ref{tab3ablation}, we draw the following observations:
\begin{itemize}
\item \textbf{Effectiveness of BERT.} Compared with BERT+BiLSTM+SFSN\\+GAIN and Glove+BiLSTM+SFSN+GAIN, the former achieves 2.85\% and 1.65\% boosts in accuracy on RumourEval and PHEME respectively. It proves the effectiveness of BERT including feature ensemble.
\item \textbf{Effectiveness of SFSN.} BERT+BiLSTM+SFSN boosts the performance as compared with BERT+BiLSTM, showing 4.70\% and 5.66\% improvements in accuracy on the two datasets, which reveals that SFSN enhancing semantic associations and fusion is effective.
\item \textbf{Effectiveness of GAIN.} Compared to BERT+BiLSTM+SFSN, BERT+BiLSTM+SFSN+GAIN notably improves the performance (2.65\% and 1.49\% boosts) on the two datasets with the help of conflicting semantics obtained by GAIN. It explains the effectiveness of GAIN.
\end{itemize}

\subsubsection{Semantic-level Fusion Self-attention Networks Evaluation}
\label{multiScaleEval}
To further evaluate the effectiveness of SFSN, we verify the performance of semantic-level fusion between four output features of GAIN and four self-attention networks i.e., four SFSN, where 'SFSN-post-word' represents only removing the fusion between the output of GAIN for word-level of posts and self-attention networks for word-level of posts. 'SFSN-post-emotion', 'SFSN-comment-word', and 'SFSN-comment-emotion' are for only removing the fusion on emotion level of posts, only removing the fusion on word level of comments, and only removing the fusion on emotion level of comments, respectively. Table \ref{tab4sfsn} presents the experimental results of these methods on RumourEval and PHEME. We have the following observations:
\begin{itemize}
\item The removal of word-level SFSNs presents lower performance than emotion-level SFSNs, which shows up to 1.27\% and 1.04\% performance degradation than emotion-level SFSNs in accuracy on RumourEval and PHEME, respectively. It reveals that semantics are more effective than emotions for fake news detection.
\item Compared with comment-level SFSNs, post-level SFSNs decrease 0.52\% and 0.64\% performance, which indicates posts contain more effective credibility indicators.
\item The removal of any SFSN will degrade the performance of the model, which shows from 2.84\% to 1.91\% degradation in accuracy on RumourEval and PHEME, respectively. This demonstrates the effectiveness of SFSN.
\end{itemize}
\begin{table}
\footnotesize
	\center
	\caption{Comparison between different SFSNs.}
	\label{tab4sfsn}
\setlength{\tabcolsep}{2.4mm}{
	\begin{tabular}{llcccc} \hline  \rule{0pt}{12pt}
        & & A(\%) & P(\%) & R(\%) & F1(\%) \\ \hline \\[-6pt]
		\multirow{5}{*}{\rotatebox{90}{RumourEval}}
        & SFSN-post-word & 80.05 & 76.87 & 84.03 & 80.29 \\
        & SFSN-post-emotion & 81.20 & 77.82 & 85.34 & 81.40 \\
        & SFSN-comment-word	& 80.57 & 77.24 & 84.69 & 80.79 \\
        & SFSN-comment-emotion & 81.84 & 78.13 & 85.94 & 81.85 \\
        & Our Model      & 82.89 & 78.52 & 86.21 & 82.19 \\ \hline  \\[-6pt]
        \multirow{5}{*}{\rotatebox{90}{PHEME}}
        & SFSN-post-word & 80.54 & 77.37 & 83.21 & 80.18 \\
        & SFSN-post-emotion & 81.51 & 78.26 & 84.34 & 81.19 \\
        & SFSN-comment-word	& 80.95 & 77.91 & 83.78 & 80.74 \\
        & SFSN-comment-emotion & 81.96 & 78.68 & 84.76 & 81.61 \\
        & Our Model      & 82.45 & 79.12 & 85.20 & 82.05 \\ \hline  \\[-6pt]
	\end{tabular}
}
\end{table}

\begin{figure}
	\centering
	\includegraphics[width=0.48\textwidth]{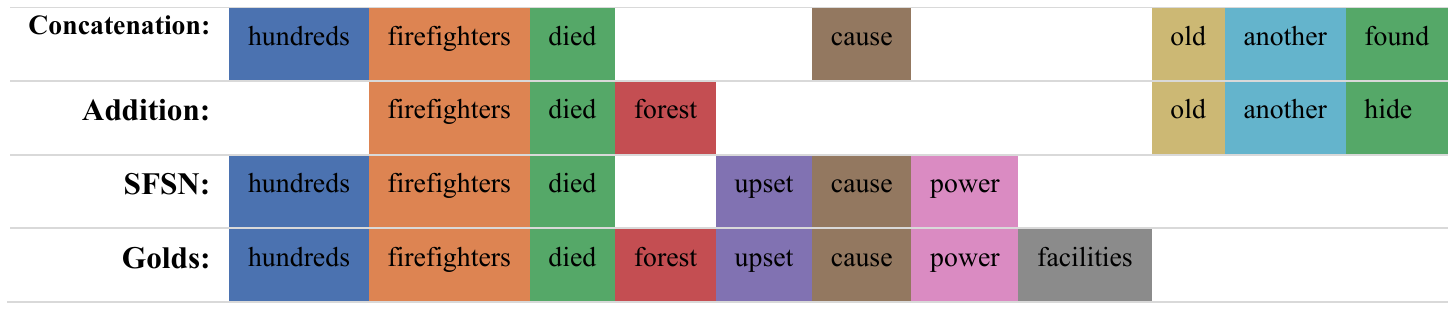}
	\caption{The words captured by SFSN, concatenation, and addition at word-level in posts.}
	\label{fig5selfAdapt}
\end{figure}

To validate the capture of valuable features in SFSN, we respectively map the outputs of self-attention networks and simple fusion strategies (which replaces SFSN) to the input elements of word-level and the visualized results are shown in Figure \ref{fig5selfAdapt} where the input is the tweet in Figure 1(b). We observe that:
\begin{itemize}
\item Concatenation and addition can capture partial keywords, like `firefighters' and `died' as well as obtaining some non-keywords, like `old' and `another', which illustrates that simple fusion strategies easily introduce noise.
\item SFSN can catch most of keywords compared with golds while it does not touch any irrelevant semantics, which confirms that SFSN is capable of effectively capturing noteworthy features.
\end{itemize}

\subsubsection{Gated Adaptive Interaction Networks Evaluation}
\label{balancedEval}
Table \ref{tab4selfAdapt} provides the performance of each part of GAIN by the following simplified models: \textbf{-All} means that GAIN is replaced by concatenation; \textbf{-Conflicting} implies that only refining gate works in GAIN; \textbf{-Refining} means that the refining gate is removed; and \textbf{-Adaptive} means that adaptive mechanism is replaced with concatenation.

From Table \ref{tab4selfAdapt}, we observe that:
\begin{itemize}
\item Ablating any parts of GAIN can reduce the performance of model, underperforming AIFN from 0.44\% to 2.65\% degradation in accuracy on RumourEval and PHEME, respectively. This confirms the effectiveness of each part of GAIN;
\item -Conflicting achieves the worst performance compared to the models that only ablate one part, showing at least 1.29\% degradation in accuracy on the two datasets, which explains that conflicting semantics obtained by -Conflicting can more strongly improve the performance of fake news detection.
\end{itemize}

\begin{table}
\footnotesize
	\center
	\caption{Comparison among different parts of GAIN.}
	\label{tab4selfAdapt}
\setlength{\tabcolsep}{3.65mm}{
	\begin{tabular}{llcccc} \hline   \rule{0pt}{12pt}
        & & A(\%) & P(\%) & R(\%) & F1(\%) \\ \hline \\[-6pt]
		\multirow{5}{*}{\rotatebox{90}{RumourEval}}
        & -All & 80.24 & 77.18 & 84.23 & 80.79 \\
        & -Conflicting & 81.04 & 77.26 & 85.48 & 81.16 \\
        & -Refining	& 81.32 & 77.74 & 84.56 & 81.01 \\
        & -Adaptive	& 82.45 & 78.03 & 85.89 & 81.77 \\
        & AIFN      & 82.89 & 78.52 & 86.21 & 82.19 \\ \hline  \\[-6pt]
        \multirow{5}{*}{\rotatebox{90}{PHEME}}
        & -All & 80.96 & 77.56 & 83.66 & 80.49 \\
        & -Conflicting & 81.16 & 77.91 & 84.21 & 80.94 \\
        & -Refining	& 81.52 & 78.53 & 84.78 & 81.54 \\
        & -Adaptive	& 81.89 & 78.87 & 84.95 & 81.80 \\
        & AIFN	    & 82.45 & 79.12 & 85.20 & 82.05 \\ \hline  \\[-6pt]
	\end{tabular}
}
\end{table}

\subsubsection{Error Analysis}
According to the results of Table 2, our model achieves unsatisfied performance in recall compared with MTL-LSTM by -2.86\% and -2.67\% on RumourEval and PHEME, respectively. Two reasons can be explained for this issue: 1) MTL-LSTM builds multi-task learning, which can learn the relationship features between tasks. Our model focuses on the interaction and interaction of multiple effective features under a single task – fake news detection, lacking joint training of multiple tasks. 2) Stance features may have significant advantages in recall rate, but have no obvious performance boosts and even bring negative effects on other evaluation metrics. In order to improve the performance of our model effectively and equitably, we do not leverage stance features to interact with other credibility indicators.

\begin{figure}
	\centering
	\includegraphics[width=0.48\textwidth]{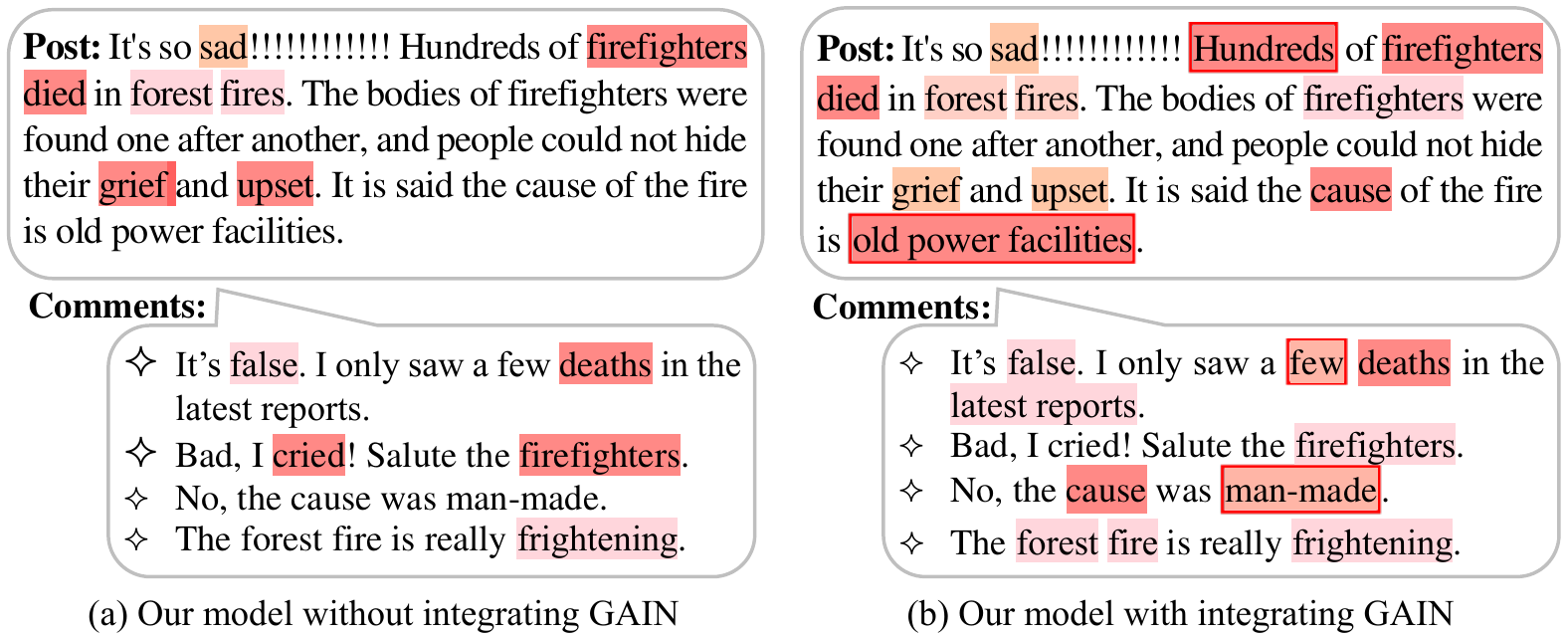}
	\caption{The visualization of fusion interaction of AIFN.}
	\label{fig6gateCase2}
\end{figure}

\subsection{Case Study}
\label{caseStudy}
To express intuitively what features AIFN has learned from posts and comments, we design two groups of experiments to visualize the outputs of semantic-level fusion self-attention networks (SFSN). The first group is SFSN integrating GAIN (our model with integrating GAIN) and the second is that SFSN integrates the outputs of concatenation which replaces GAIN (our model without integrating GAIN). Specifically, we firstly employ max-pooling operation to pool the outputs of four self-attention networks and then map them into the corresponding elements in the input layer respectively, and finally obtain interesting patterns to visualize in Figure \ref{fig6gateCase2}. We observe that:

\begin{itemize}
\item From Figure \ref{fig6gateCase2} (a), our model without integrating GAIN not only obtains some keywords, like `firefighters' and `died' but also highlights semantic associations between posts and comments, such as `died' to `deaths', `grief, upset' to `cried', and `firefighters', which gets benefit from SFSN screening and feature correlation.
\item From Figure \ref{fig6gateCase2} (b), the model with GAIN captures similar semantics between posts and comments, like `died' and `cause', and also focuses on some conflicting words, like `hundred', `old power facilities' in posts and `few', `man-made' in comments. It reveals GAIN effectively finds conflicting semantics about fake news from comments.
\end{itemize}

\section{Conclusion}
\label{sec5conclusion}
In this paper, we studied the problem of cross-interaction fusion in fake news detection on social media and proposed adaptive interaction fusion networks considering the extracted features from posts and comments. Specifically, we discovered conflicting semantics between posts and comments by gated adaptive interaction networks and developed semantic-level fusion self-attention networks to explore feature associations for capturing noteworthy features and fusing deeply them. The experimental results based on two real-world datasets demonstrated that our method significantly outperformed previous state-of-the-art models. In the future, we will capture semantic conflicts by considering the hierarchical interaction structure of comments for fake news detection.

\ack We would like to thank the referees for their comments. The research work is supported by ``National key research and development program in China" (2019YFB2102300), ``the World-Class Universities(Disciplines) and the Characteristic Development Guidance Funds for the Central Universities" (PY3A022), Ministry of Education Fund Projects (No. 18JZD022 and 2017B00030), Shenzhen Science and Technology Project (JCYJ20180306170836595), Basic Scientific Research Operating Expenses of Central Universities (No.ZDYF2017006), Xi'an Navinfo Corp.\& Engineering Center of Xi'an Intelligence Spatial-temporal Data Analysis Project (C2020103); Beilin District of Xi'an Science \& Technology Project (GX1803).

\bibliography{ecai}

\end{document}